\documentclass[journal]{IEEEtran}
\usepackage{multirow}

\usepackage{hyperref}
\usepackage{graphicx,subfigure}
\usepackage{subfigure}
\usepackage{amsmath,amssymb}
\usepackage{algorithm}
\usepackage{algorithmic}
\usepackage{rotating}
\usepackage[switch,mathlines]{lineno}

\ifCLASSINFOpdf
\else
\fi

\begin{document}
%
\title{Data-Driven Scene Understanding with Adaptively Retrieved Exemplars}


\author{Xionghao~Liu,
        Wei~Yang,
        Liang~Lin,
        Qing~Wang,
        Zhaoquan~Cai,
        and~Jianhuang~Lai,~\IEEEmembership{Member,~IEEE}
        \thanks{This work was supported by the National Natural Science Foundation of China (no. 61170193, no. 61370185), Guangdong Science and Technology Program (no. 2012B031500006), Guangdong Natural Science Foundation (no. S2012020011081), Special Project on Integration of Industry, Education and Research of Guangdong Province (no. 2012B091100148, no. 2012B091000101). This work is partially supported by the Hong Kong Scholar program.}
        \thanks{X. Liu, W. Yang, Q. Wang, L. Lin, J. Lai are with Sun Yat-sen University, Guangzhou 510006, China. Z. Cai is with Huizhou University, Huizhou, China. L. Lin is also with the Department of Computing, The Hong Kong Polytechnic University, Hong Kong, China. e-mail: (ericwangqing@gmail.com).}}

\markboth{IEEE Multimedia, 2015}%
{X. Liu, \MakeLowercase{\textit{et al.}}: Data-Driven Scene Understanding with Adaptively Retrieved Exemplars}
%



\IEEEtitleabstractindextext{%
\begin{abstract}
This article investigates a data-driven approach for semantically scene understanding, without pixelwise annotation and classifier training. Our framework parses a target image with two steps: (i) retrieving its exemplars (i.e. references) from an image database, where all images are unsegmented but annotated with tags; (ii) recovering its pixel labels by propagating semantics from the references. We present a novel framework making the two steps mutually conditional and bootstrapped under the probabilistic Expectation-Maximization (EM) formulation. In the first step, the references are selected by jointly matching their appearances with the target as well as the semantics (i.e. the assigned labels of the target and the references).
We process the second step via a combinatorial graphical representation, in which the vertices are superpixels extracted from the target and its selected references. Then we derive the potentials of assigning labels to one vertex of the target, which depend upon the graph edges that connect the vertex to its spatial neighbors of the target and to its similar vertices of the references. 
Besides, the proposed framework can be naturally applied to perform image annotation on new test images. In the experiments, we validate our approach on two public databases, and demonstrate superior performances over the state-of-the-art methods in both semantic segmentation and image annotation tasks.

\end{abstract}

\begin{IEEEkeywords}
scene understanding, semantic segmentation, image retrieval, graphical model, image annotation
\end{IEEEkeywords}}

\maketitle

\IEEEdisplaynontitleabstractindextext

%
\IEEEpeerreviewmaketitle

\begin{figure}[tb]
\begin{center}
\includegraphics[width=0.95\linewidth]{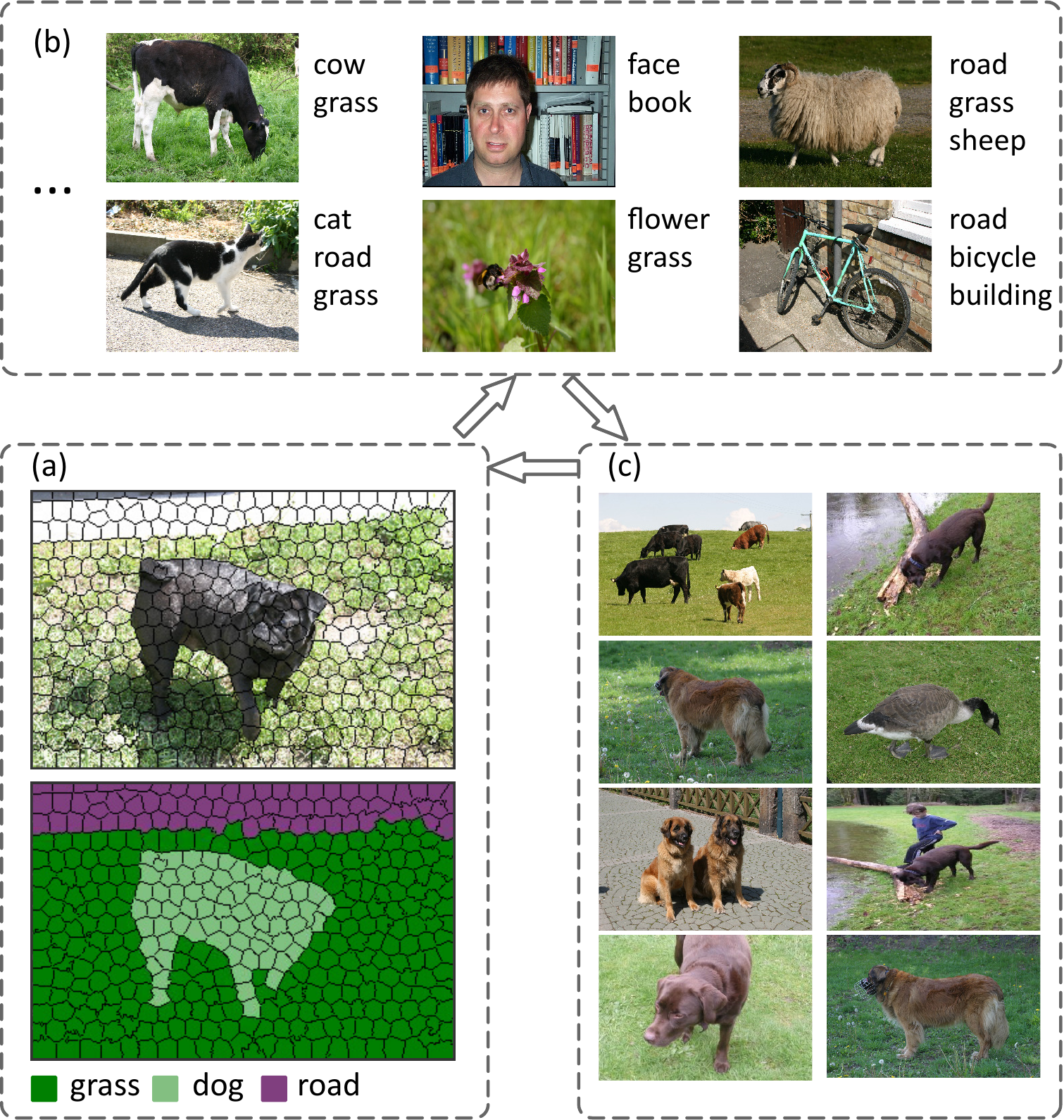}
\end{center}
   \caption{A glance of our framework, where we semantically segment the target image in a self-driven fashion: The algorithm iterates to retrieve (c) the exemplars matching with the target from (b) the auxiliary data , and (a) parse the target image in the virtue of the strength of the selected exemplars.}
\label{fig:framework}
\end{figure}

\section{Introduction}


\IEEEPARstart{S}ignificant progresses have been identified in solving the task of semantic image understanding~\cite{shotton2006textonboost,ladicky2010graph}. However, these methods usually build upon supervised learning with fully annotated data that are expensive and sometimes limited in large-scale scenarios~\cite{lin2009grammar,lin2012representing}. Several weakly supervised methods were proposed~\cite{zhang2013sparse} to reduce the overload of data annotating, which can be trained with only image-level labels indicating the classes presented in the images. Recently, data-driven approaches \cite{liu2009nonparametric,luo2012joint} receive increasing attentions, which tend to leverage knowledges from auxiliary data in weakly supervised fashions, and demonstrate very promising applications. Following this trend, one interesting but challenging problem arises for the scene understanding: How to parse the raw images in virtue of the strength of numerous unsegmented but tagged images, as the image-level tags can be achieved easier. 
In this work, we investigate this problem by developing a unified framework, in which the two following steps perform iteratively, as Fig. \ref{fig:framework} illustrates.

In {\bf Step. 1}, we search for similar images as the exemplars (i.e. references) matching to the target image from the auxiliary database (in Fig. 1 (b)), and these references are required to share similar semantic concepts with the target. 
Moreover, we enforce the representation to be semantically meaningful: The references that are selected should contain consistent tags. The tags of the target image can be also taken into account during the iteration, as they can be determined by the last label assignment step (in Step. 2).  We solve this step using the proximal gradient method. 


In {\bf Step. 2}, we assign labels to the pixels of the target by propagating semantics from the selected references.
We create a graphical model, in which the vertices are the superpixels from the target image and its references. There are two types of edges defined over the graph, which is inspired by~\cite{lin2010layered}: (i) the inner-edges connecting the spatial adjacent vertices within the target; (ii) the outer-edges connecting the vertices of the target to those of its references.  The potentials are then derived into an MRF form by aggregating the two types of edge connections, which can be fast solved by the Graph Cuts algorithm \cite{ladicky2010graph}. 

The two above steps are mutually conditional, providing complementary information to each other. We present a novel probabilistic Expectation-Maxima (EM) formulation making the two steps bootstrapped by each other to conduct results in a self-driven manner. In addition, the proposed framework can also be directly applied on new test image to perform multi-label image annotation. Our approach is  evaluated on several benchmarks, and outperforms other state-of-the-art methods. 


\section{Related work}

Traditional efforts for scene understanding mainly focused on capturing scene appearances, structures and spatial contexts by developing combinatorial models, e.g., CRF \cite{shotton2006textonboost, ladicky2010graph}, Texton-Forest \cite{shotton2008semantic}, Graph Grammar~\cite{lin2014AOG}.  These models were generally founded on supervised learning techniques, and required manually prepared training data containing labels at pixel level.

Several weakly supervised methods are proposed to indicate the classes presented in the images with only image-level labels. For example, Winn et al.~\cite{winn2005locus} proposed to learn object classes based on unsupervised image segmentation. Zhang et al. \cite{zhang2013sparse} learned classification models for all scene labels by selecting representative training samples, and multiple instance learning was utilized in \cite{vezhnevets2011weakly}. 

Some nonparametric approaches have been also studied that solve the problems by searching and matching with an auxiliary image database. For example, an efficient structure-aware matching algorithm was discussed in \cite{liu2009nonparametric} to transfer labels from the database to the target image, but the pixelwise annotation was required for the auxiliary images. 


\section{Problem Formulation} \label{sec_formulation}
In this section, we phrase the problem in a probabilistic formulation, and then discuss the Expectation-Maximization (EM) inference framework for optimization.

\subsection{Probability Model}

Let $\varDelta = \{I_k, L_k\}_{k=1}^N$ denote a set of images $\{I_k\}$ with image-level labels $\{L_k\}$. Each image $I _k $ is represented as a set of superpixels $\{x^k_i\}_{i=1}^{n_{k}}$, where $n_k$ is the number of superpixels in $I_k$.


Given the target image $I_t$, our task is to predict its image-level labels $L_t$, as well as to assign each superpixel $x^t_i$  a label $y^t_i \in L_t$. Let $Y_t$  denote the whole label assignment, \emph{i.e.}, $Y_t = \{y^t_i\}_{i=1}^{n_t}$, we can define the joint probability distribution of target image $I_t$ and the label assignment $Y_t$.

We also define a binary-valued correspondence variable $\boldsymbol{\alpha} = \{\alpha_k\}_{k=1}^{N}$ such that $\alpha_k=1$ if image $I_k$ is selected as a reference for the target image. $\boldsymbol{\alpha}$ is treated as a hidden variable.


The complete probability model is defined as follow,
\begin{equation}
  P(I_t, Y_t, \boldsymbol{\alpha} | \varDelta) = P(I_t, Y_t | \boldsymbol{\alpha}, \varDelta) P(\boldsymbol{\alpha}),
\label{eq:model}
\end{equation}
and we further derive it by summing out $\boldsymbol{\alpha}$ as,
\begin{equation}
  P(I_t, Y_t| \varDelta) = \sum_{\boldsymbol{\alpha}} P(I_t, Y_t | \boldsymbol{\alpha}, \varDelta) P(\boldsymbol{\alpha}).
\label{eq:model_summation}
\end{equation}
Then the optimal label assignment $Y_t$ by maximizing the probability,
\begin{equation}
  Y_t^* = \arg\max_{Y_t} P(I_t, Y_t| \varDelta),
\label{eq:target}
\end{equation}
and we propose to solve it iteratively under an Expectation-Maximization (EM) framework.

\begin{figure}[htp]
\begin{center}
\includegraphics[width=0.8\linewidth]{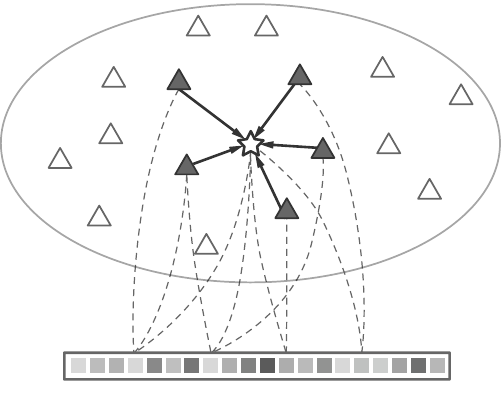}
\end{center}
   \caption{Illustration of the semantic-aware sparse coding. \textbf{Top:} The target image is denoted by the pentagon and each auxiliary image denoted by an triangle. The darked triangles represent the images selected as the references. \textbf{bottom:} The grey squares represent semantic labels that are introduced as constraints during the optimization.  And we select a subset of auxiliary images as references for the target image.}
\label{fig:label-assignment}
\end{figure}
\subsection{The EM Iterations}

It has been shown that estimating $ Y_t^*$ from $ P(I_t, Y_t| \varDelta)$  is equivalent to minimize the following energy function~\cite{neal1998view}:
\begin{equation} 
\label{eq:target-energy}
  \mathcal{L}(Q, Y_t) =
  -\sum_{\boldsymbol{\alpha}} Q(\boldsymbol{\alpha})\ln P(I_t, Y_t, \boldsymbol{\alpha} | \varDelta)
   + \sum_{\boldsymbol{\alpha}} Q(\boldsymbol{\alpha}) \ln Q(\boldsymbol{\alpha}),
\end{equation}
where $Q(\boldsymbol{\alpha})$ is the posterior of the latent variable $\boldsymbol{\alpha}$.

Since the second term in Eq. (\ref{eq:target-energy}) is a constant, the optimization  iterates with two steps:  (i) The \emph{E-step} minimizes the energy $\mathcal{L}(Q, Y_t)$ with respect to $Q(\boldsymbol{\alpha})$  with $Y_t$ fixed. 
(ii) The \emph{M-step} minimizes the energy $\mathcal{L}(Q, Y_t)$ with respect to $Y_t$ with $Q(\boldsymbol{\alpha})$ fixed.

\vspace{0.4em}\noindent\textbf{(i) The E-step: Approximating $Q(\boldsymbol{\alpha})$ :}
\label{sec_sasc}

The posterior of the latent variable $Q(\boldsymbol{\alpha})$ is defined as,
\begin{equation}
  Q(\boldsymbol{\alpha}) = P(\boldsymbol{\alpha}|I_t, Y_{t}, \varDelta) = \frac{1}{Z}\exp\{-E_{\boldsymbol{\alpha}}(\boldsymbol{\alpha}, I_t, Y_t, \varDelta)\},
  \label{eq:estep}
\end{equation}
where $Z$ is the normalization constant of the probability. The energy $E_{\boldsymbol{\alpha}}$ 
evaluates the appearance and semantics consistency, which is specified as,
\begin{equation}\label{eq:energy-alpha}
E_{\boldsymbol{\alpha}}(\boldsymbol{\alpha}, I_t, Y_t, \varDelta) = E_{Sc}(\boldsymbol{\alpha}, I_t, \varDelta)+ \gamma \ E_{Sa}(\boldsymbol{\alpha}, Y_{t}, \varDelta),
\end{equation}

The first term $E_{Sc}$  measures the appearance similarity between $I_t$ and images in $\varDelta$,  defined as,
\begin{equation}
  E_{Sc} = \frac{1}{2}\|F(I_t)-B\boldsymbol{\alpha}\|_2^2+ \beta \|\boldsymbol{\alpha}\|_1,
   \label{eq_E_sc}
\end{equation}
where $\beta$ is the tradeoff parameter used to balance the sparsity and the reconstruction error. $F(\cdot)$ is an $m$-dimemsional global feature of an image, and $B\in \mathbb{R}^{m \times N}$ is a matrix consisting of all the features of images in $\varDelta$.

The second term $E_{Sa}$ in Eq. (\ref{eq:energy-alpha}) measures semantic consistency, defined as,
\begin{eqnarray}
  E_{Sa} & = & \frac{1}{2} \sum_{i, j \in N} \mathcal{S}_{ij} \|\frac{\alpha_i}{\sqrt{A_{ii}}}-\frac{\alpha_j}{\sqrt{A_{jj}}}\|_2^2  + \lambda \ \boldsymbol{\alpha}^T \mathcal{D} \boldsymbol{\alpha} \nonumber \\
  & = & \boldsymbol{\alpha}^T \mathcal{L} \boldsymbol{\alpha} + \lambda \ \boldsymbol{\alpha}^T \mathcal{D} \boldsymbol{\alpha},
  \label{eq_E_sa}
\end{eqnarray}
where $\mathcal{S}_{ij}$ measures the semantic similarity between $(I_i, I_j) \in \varDelta$, as,
\begin{equation}
  \mathcal{S}_{ij}=\frac{|L_i \cap L_j|}{|L_i \cup L_j|}.
\label{eq_w}
\end{equation}
and $A$ in Eq. (\ref{eq_E_sa})  is a diagonal matrix where $A_{ii}=\sum_{j=1}^{N}\mathcal{S}_{ij}$ and $\mathcal{L}=A^{-1/2}(A-\mathcal{S})A^{-1/2}$, in which $L$ is the normalized Laplacian matrix.

Images with similar semantics should be encoded with similar activations. In other words, if two images have common labels, then the activations corresponding to this image pair should also be close to each other. The distance between their activation codes should be small.

$\mathcal{D}$ is a diagonal matrix where $\mathcal{D}_{kk}$ measures the semantic dissimilarity between $I_k \in \varDelta$ and the target image $I_t$. Thus the second term\footnote{$\boldsymbol{\alpha}^T \mathcal{D} \boldsymbol{\alpha}$ is convex, and it is convenience for optimization.} $\boldsymbol{\alpha}^T \mathcal{D} \boldsymbol{\alpha}$ penalizing the target $I_t$ is reconstructed by images that are semantically dissimilar with $I_t$. We define the diagonal matrix $\mathcal{D}$ by
\begin{equation}
\mathcal{D}_{kk}=1-\frac{|L_t \cap L_k|}{|L_t \cup L_k|},
\label{eq_D}
\end{equation}
where $L_t$ are the latent labels of the target image, which are unknown at the beginning\footnote{We initialize $L_t$ as the whole label set of the database.}, and can be determined from $Y_t$ during the later iterations.

\vspace{0.4em}\noindent\textbf{(ii) The M-step: estimating $Y_t$ :}\label{sec_crf}

The M-step performs to minimize the following energy function with respect to $Y_t$:
\begin{equation}
\label{eq:mstep}
E_M(Y_t) = - \sum_{\boldsymbol{\alpha}} Q(\boldsymbol{\alpha})\ln P(I_t, Y_t, \boldsymbol{\alpha} | \varDelta).
\end{equation}	
However, summing out $\boldsymbol{\alpha}$ for all possibilities demands very expensive computational cost, particularly to process a large number $N$ of data. Instead, we seek a lower-bound of $E_M(Y_t)$. Assume that we can infer $\boldsymbol{\alpha}^*$ with the maximized probability $Q(\boldsymbol{\alpha}^*)$ by the E-step. Then we can define the joint distribution of $(I_t, Y_t)$ conditioned on $Q(\boldsymbol{\alpha}^*)$, and we have
\begin{equation}
  \sum_{\boldsymbol{\alpha}} P(I_t, Y_t  | \varDelta; \boldsymbol{\alpha}^*  ) >  \sum_{\boldsymbol{\alpha}} P(I_t, Y_t, \boldsymbol{\alpha} | \varDelta).
\end{equation}	
It is straightforward in the context of our task, as the cumulative density of assigning labels from good references (\emph{i.e.} given $\boldsymbol{\alpha}^*$) is higher than that with general cases. Thus, we set the lower-bound as,
\begin{equation}
E_M(Y_t) >  - \sum_{\boldsymbol{\alpha}} Q(\boldsymbol{\alpha}) \ln P(I_t, Y_t,  | \varDelta; \boldsymbol{\alpha}^*  ) ,
\end{equation}	
where $Q(\boldsymbol{\alpha})$ is fixed by the last E-step.  The energy to be minimized can be further simplified as,
\begin{equation}
\label{eq:mstep_star}
\hat{E_M}(Y_t) = - \ln P(I_t, Y_t | \varDelta, \boldsymbol{\alpha}^* ),
\end{equation}
where we will specify $-\ln P(I_t, Y_t  | \varDelta, \boldsymbol{\alpha}^*)$ with a combinatorial graph model in Sec. \ref{sec:label-assignment}.

\section{Inference and Implementation} \label{sec_infer}

Within the EM formulation, the inference algorithm iterates with two steps: (i) computing $\boldsymbol{\alpha}^*$ in the E-step for reference retrieval and (ii) solving the optimal labeling $Y_t^*$ with the selected references in the M-step.

\begin{algorithm}[htpb]
\caption{Adaptive Reference Retrieval}
\label{alg:Ea}
\textbf{Input:} Target image feature $F(I_t)$, codebook $B$ , semantic constrains $\varLambda$, and the threshold $\sigma$ for stop.\\
\textbf{Output:}  Semantical sparse coding coefficient $\boldsymbol{\alpha^*}$.\\
\textbf{Initial:} Initial $\boldsymbol{\alpha}^*$ in randomly , and $k=1$. Denote $g(\boldsymbol{\alpha})=\frac{1}{2} \|F(I_t)-B\boldsymbol{\alpha}\|_2 + \frac{1}{2} \gamma \  \boldsymbol{\alpha}^T \varLambda \boldsymbol{\alpha}$,
so Eq. (\ref{eq_combine}) can be reformulated as $E_{\alpha} = g(\boldsymbol{\alpha})+ \beta \|\boldsymbol{\alpha}\|_1$.

\begin{algorithmic}[1]
\WHILE{$\|\boldsymbol{\alpha}^{k+1}-\boldsymbol{\alpha}^{k}\|_2 > \sigma$}
    \STATE Compute the gradient of $g(\boldsymbol{\alpha})$ at $\boldsymbol{\alpha}^k$,
    $\triangledown g(\boldsymbol{\alpha}^k) = B^T(B\boldsymbol{\alpha}^k-F(I_t)) + \gamma \varLambda \boldsymbol{\alpha}^k$.
    \STATE
    \label{alg_step_opt}
    $\boldsymbol{z}^*_L=\arg\min_{\boldsymbol{z}}(\boldsymbol{z}-\boldsymbol{\alpha}^k)^T
    \triangledown g(\boldsymbol{\alpha}^k) + \beta \|\boldsymbol{z}\|_1 + \frac{L}{2} \|\boldsymbol{z}-\boldsymbol{\alpha}^k\|_2$,
    where $L>0$ is a papameter.
    \STATE Iteratively increasing $L$ by a constant factor until the condition
    $g(\boldsymbol{z}^*_L) \leq M_g^L(\boldsymbol{\alpha}^k,\boldsymbol{z}_L^*) :=
    g(\boldsymbol{\alpha}^k)+\triangledown g(\boldsymbol{\alpha}^k)^T(\boldsymbol{z}_L^*-\boldsymbol{\alpha}^k)
    + \frac{L}{2} \|\boldsymbol{z}_L^* - \boldsymbol{\alpha}^k \|_2$ is met, else return to step \ref{alg_step_opt}.
    \STATE Update
    $\boldsymbol{\alpha}^{k+1} := \boldsymbol{\alpha}^{k} + \nu_k(\boldsymbol{z}^*_L - \boldsymbol{\alpha}^k)$, where $\nu_k \in (0,1]$
    \STATE k:=k+1
\ENDWHILE
\STATE $\boldsymbol{\alpha}^* = \boldsymbol{\alpha}^k$
\end{algorithmic}
\end{algorithm}

\subsection{Adaptive Reference Retrieval}\label{sec:reference-retrieval}

Maximizing  $Q(\boldsymbol{\alpha})$ is equivalent to minimizing the energy defined in  Eq. (\ref{eq:energy-alpha}) \emph{w.r.t.} $\boldsymbol{\alpha}^* = \arg \min_{\boldsymbol{\alpha}} E_{\boldsymbol{\alpha}}(\boldsymbol{\alpha}, I_t, Y_t, \varDelta)$.
Notice that $E_{\boldsymbol{\alpha}}(\boldsymbol{\alpha}, I_t, Y_t, \varDelta)$ can be regarded as a semantic-aware
sparse representation, where we jointly model the appearance reconstruction with semantic consistency. Fig. \ref{fig:label-assignment} intuitively illustrates this model, and it can be rewritten as,
\begin{equation}
E_{\alpha} = \frac{1}{2} \|F(I_t)-B\boldsymbol{\alpha}\|_2 +
\beta \|\boldsymbol{\alpha}\|_1 +
\frac{1}{2} \gamma \  \boldsymbol{\alpha}^T \varLambda \boldsymbol{\alpha},
\label{eq_combine}
\end{equation}
where $\varLambda = 2(\mathcal{L}+\lambda \mathcal{D})$.
The semantic associated terms in Eq. (\ref{eq_combine}) can be phrased in convex forms, thus we can use the proximal gradient method to solve this problem efficiently. The optimization process is shown in Algorithm \ref{alg:Ea}.

Given the optimized $\boldsymbol{\alpha}^*$, we can simply select the references according to coding co-efficiencies, e.g., select by thresholding. And we set $\alpha_k=0$ if image $I_k$ is not selected.



\begin{figure}[htp]
\begin{center}
\includegraphics[width=0.8\linewidth]{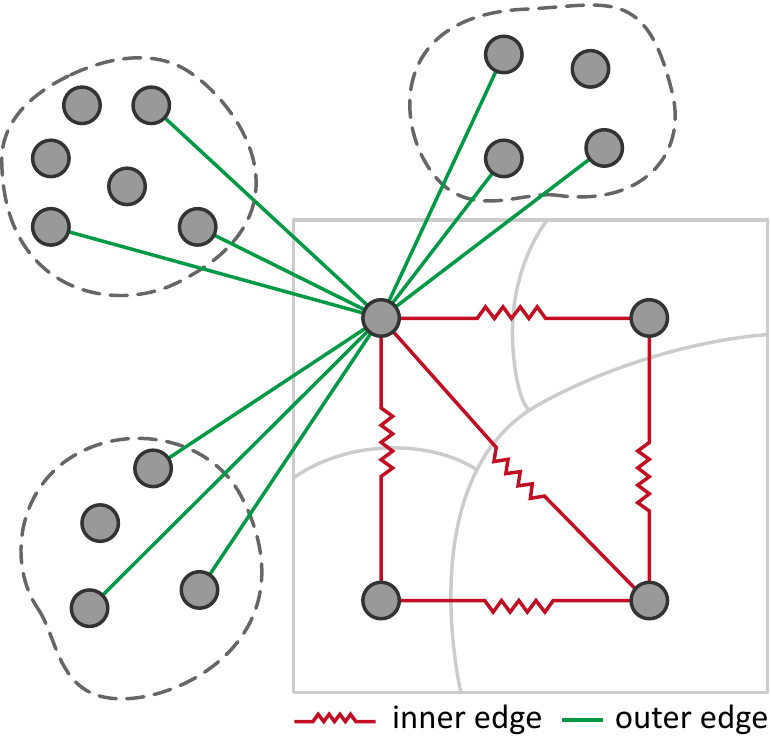}
\end{center}
\caption{Illustration of the combinatorial graphical model. The dark circles represent the superpixels; the fours over the square region are extracted from the target image while the others from references that are denoted by dashed regions.}  
\label{fig:graphical-model}
\end{figure}

\subsection{Aggregated Label Assignment}\label{sec:label-assignment}

Given the references determined by $\boldsymbol{\alpha}^*$, we propagate their semantic labels to $I_t$ by constructing a combinatorial graph. We extract superpixels from both $I_t$ and the references as graph vertices, and connect them with probabilistic edges incorporating their affinities, as Fig.~\ref{fig:graphical-model} illustrates.

Two types of edges are considered over the graph: (i) the inner-edges $\omega$ connecting the spatial neighboring superpixles within the target (red wavy line in Fig. \ref{fig:graphical-model}) , and (ii) the outer-edges $\xi$ connecting the superpixels of the target to those of its references (straight green line in Fig. \ref{fig:graphical-model}) . And each superpixel of the target connects with the $q$ most similar superpixels of each reference.

We define $-\ln P(I_t,Y_t|\varDelta,\boldsymbol{\alpha}^*)$ in Eq. (\ref{eq:mstep_star}) on the graphical model as follows,
\begin{eqnarray}
\label{eq:mstep-simplified}
-\ln P(I_t,Y_t|\varDelta,\boldsymbol{\alpha}^*) = &
\sum_{i=1}^{n_t} \psi(y_i^t|\boldsymbol{\alpha}^*, \varDelta) +
\\  \nonumber & \sum_{(x_i^t,x_j^t) \in \omega } \phi(y_i^t,y_j^t,x_i^t,x_j^t)
\end{eqnarray}
where $\omega$ is the inner edges. 
The optimization of  Eq. (\ref{eq:mstep_star}) becomes a tractable graphical model optimization problem. 

To derive the potentials of assigning labels to one vertex of the target $\psi(y_i^t|\boldsymbol{\alpha^*}, \varDelta)$ in Eq. (\ref{eq:mstep-simplified}), we propose the \emph{semantic-based superpixel density prior}, which is defined as,
\begin{equation}
  \psi(y_i^t|\boldsymbol{\alpha^*}, \varDelta) = \sum_{k=1}^{N} \alpha^*_k \rho(x_i^t,I_k) \delta(y_i^t \in L^k),
  \label{eq:data_term}
\end{equation}
where $\rho(x_i^t,I_k)$ denotes the density of superpixel $x_i^t$ in image $I_k$, which is defined as,
\begin{equation}
\rho(x_i^t,I_k) = \frac{1}{N_{\xi}}\sum_{(x_i^t,x_j^k) \in \xi} \|f(x_i^t)-f(x_j^k)\|_2,
\label{eq:density}
\end{equation}
where $\xi$  denotes outer-edges, $N_{\xi}$ is the number of outer-edges, and $f(\cdot)$ is the feature vector of a superpixel.
This density measures the similarity between the superpixel $x_i^t$ in the target and its neighboring superpixels connected by outer-edges in the reference image $I_k$, thus it implicitly exhibits the probability that $x_i^t$ sharing the same labels with its reference $I_k$.



\begin{algorithm}[htpb]
\caption{Overall procedure of our framework}
\label{alg:overall}
\textbf{Input:} Target $I_t=\{x_i^t\}_{i=1}^{n_t}$, and auxiliary $\varDelta=\{I_k,L_k\}_{k=1}^N$. \\
\textbf{Output:}  Label of each superpixel $Y_t=\{y_i^t\}_{i=1}^{n_t}$ \\
\textbf{Initial:} $L_t^1$ contains all labels, and $n=1$.

\begin{algorithmic}[1]
\WHILE{$L_t^{n+1}\neq L_t^{n}$}
    \STATE Minimize $E_\alpha$ defined in Eq. (\ref{eq_combine}) using Alg. \ref{alg:Ea}.
    \STATE Sort $\boldsymbol{\alpha^*}$ in descend order, select the images correspoding to the $p$-first nonzero coefficients, as a set $B$.
    \FORALL{$x_i^t$ in $I_t$}
        \FORALL{image $I_k$ in $B$}
        \STATE Select the $q$-most similar superpixels $O_{x_i^t}^k=\{x_j^k\}_{j=1}^q$.
        \STATE Construct $O_{x_i^t}=\cup_k O_{x_i^t}^k$
        \ENDFOR
    \STATE Add $(x_i^t,x_j^k)$ to $\omega$ for all $x_j^k \in O_{x_i^t}$.
    \STATE Add $(x_i^t,x_j^t)$ to $\xi$ for all neighbors $\{x_j^t\}$ of $x_i^t$, $i \neq j$.
    \ENDFOR

    \STATE Minimize Eq. (\ref{eq:mstep-simplified}). Optimize the latent label ${Y_t}^*$ using  alpha-beta swap algorithms of graph cuts.
\STATE Update $L_t^{n+1}$ as the unique set of $Y_t^*$.
\STATE n:= n+1

\ENDWHILE
\end{algorithmic}
\end{algorithm}

The pairwise potentials, \emph{i.e.}  $\phi(y_i^t,y_j^t,x_i^t,x_j^t)$  in  Eq. (\ref{eq:mstep-simplified}), encourages the smoothness between neighboring superpixels within the target, as,
\begin{equation}
  \phi(y_i^t,y_j^t,x_i^t,x_j^t) = \|f(x_i^t)-f(x_j^t)\|_2\delta(y_i^t \neq y_j^t),
\label{eq_pairwise_n}
\end{equation}
where $\delta(\cdot)$ is the indicator function.

Thus the approximate solutions Eq. (\ref{eq:mstep-simplified}) can be found using  alpha-beta swap algorithms of graph cuts. The sketch of our framework is shown in Algorithm \ref{alg:overall}.

\subsection{Image Annotation}
We propose a simple method to transfer $n$ labels to a test image $I_t$ from the  query's $K$ nearest neighbors in the training set. 
For a given test image $I_t$, the sparse reconstruction coefficient vector $\boldsymbol{\alpha}$ is determined by soloving the problem in Eq. (\ref{eq_combine}), where we set $\lambda=0$, and set other parameters as the same as described in section \ref{sec_pars}. The optimal sparse coefficient solution denote as $\hat{\boldsymbol{\alpha}}$, then let its top $K$ largest value denote as $\hat{\boldsymbol{\pi}} \in \Re^{K\times 1}$ consponding with image label indicator $\boldsymbol{l}_i\in \Re^C, ~ i=1,2,\dots,K$. The label vector probability of test image can then be obtained as:
\begin{equation}
\boldsymbol{z}_t = \sum_{i=1}^{K} \hat{\boldsymbol{\pi}}_i \boldsymbol{l}_i
\end{equation}
where $\hat{\boldsymbol{\pi}}_i$ is the $i$-th component of vector $\hat{\boldsymbol{\pi}}$.
The labels corresponding to the top few largest values in $\boldsymbol{z}_t$ are considered as the final annotationns of the test image.

We compare the following two annotation methods, and find out that the sparse coefficient $\boldsymbol{\alpha}$ is extremely useful for image annotation. 
(i) \textbf{weighed:} That is the annotation weighed by sparse reconstruction coefficient $\hat{\boldsymbol{\pi}}_i$.
(ii) \textbf{unweighed:} We set $\hat{\boldsymbol{\pi}}_i=1,i=1,\cdots,K$ in manual.

Besides, we also compared with classical works for image annotation, the proposed method here have the following characteristics: (i) the propagation process is robust and less sensitive to the image noises owing to the semantic constraints in image retrieval step. (ii) the proposed algorithm is scalable to large-scale, and retrieval images by jointly matching their appearances as well as the semantics.

%

\begin{figure*}[htp]
\begin{center}
\subfigure[]{\label{fig:segment result} \includegraphics[width=1\linewidth]{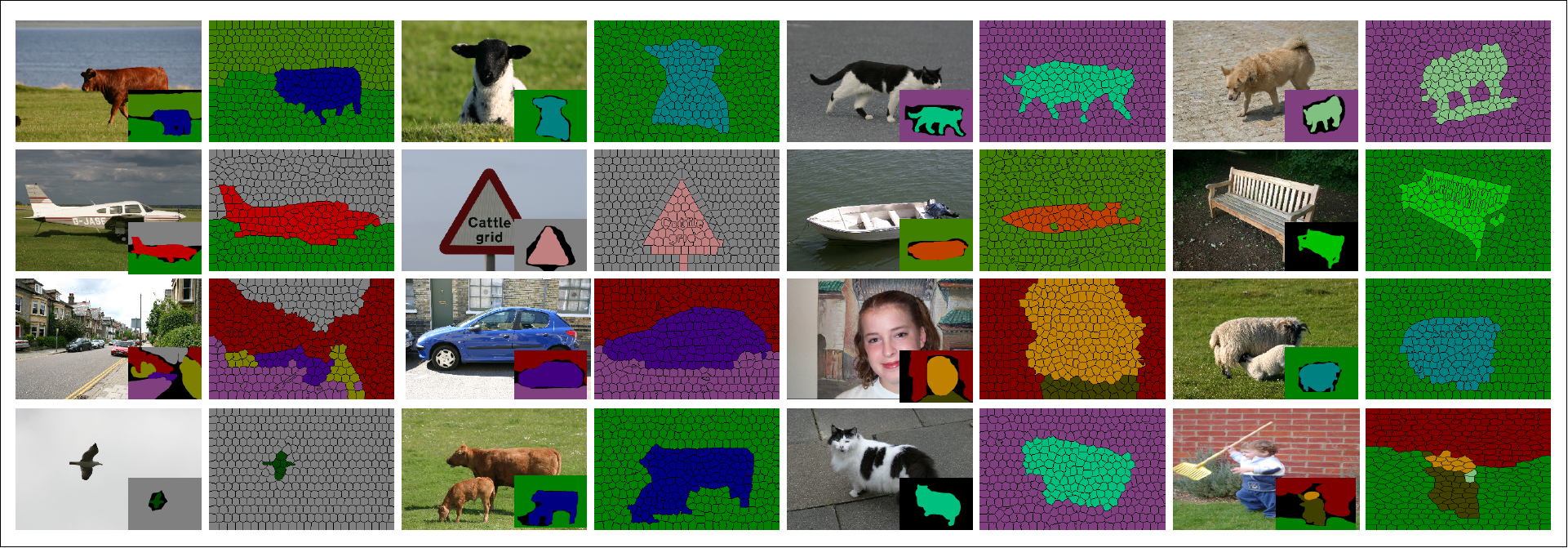}}
\subfigure[]{\label{fig:iteration_result} \includegraphics[width=1\linewidth]{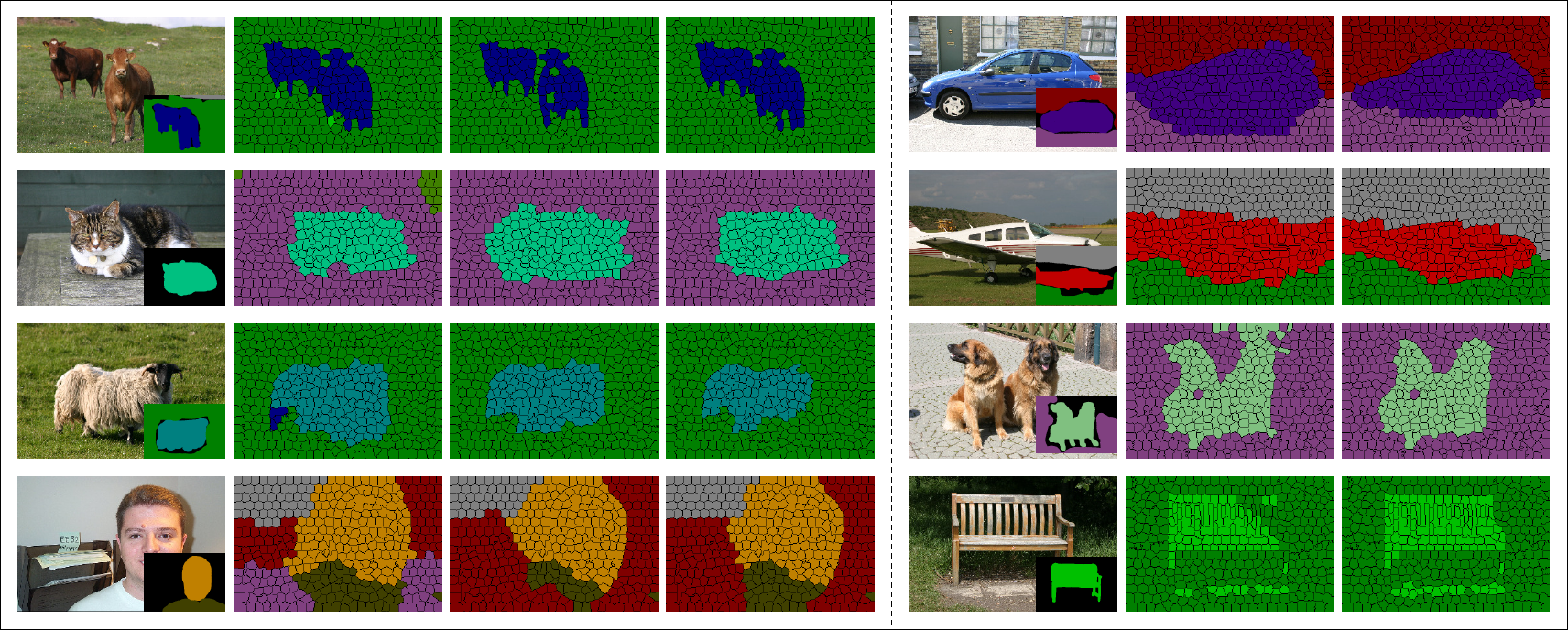}}
\end{center}
\vspace{-5mm}
\caption{Some final results (a) and some intermediate results of semantic segmentation (b) on the MSRC  dataset. The original image and its ground truth are shown on the left, and the semantic segmentation result by our method is on the right. It's encouraged to be view in color.}
\end{figure*}


\begin{table*}[htp]\footnotesize
\begin{center}
\begin{tabular}{|c|c|c|c|c|c|c|c|c|c|c|c|c|c|c|c|c|c|c|c|c|c|c|}
\hline
   \multicolumn{23}{| c |}{\textbf{MSRC}} \\
   \hline
   \normalsize{Method}
   & \begin{sideways}{building}\end{sideways}
   & \begin{sideways}{grass}\end{sideways}
   & \begin{sideways}{tree}\end{sideways}
   & \begin{sideways}{cow}\end{sideways}
   & \begin{sideways}{sheep}\end{sideways}
   & \begin{sideways}{sky}\end{sideways}
   & \begin{sideways}{airplane}\end{sideways}
   & \begin{sideways}{water}\end{sideways}
   & \begin{sideways}{face}\end{sideways}
   & \begin{sideways}{car}\end{sideways}
   & \begin{sideways}{bicycle}\end{sideways}
   & \begin{sideways}{flower}\end{sideways}
   & \begin{sideways}{sign}\end{sideways}
   & \begin{sideways}{bird}\end{sideways}
   & \begin{sideways}{book}\end{sideways}
   & \begin{sideways}{chair}\end{sideways}
   & \begin{sideways}{road}\end{sideways}
   & \begin{sideways}{cat}\end{sideways}
   & \begin{sideways}{dog}\end{sideways}
   & \begin{sideways}{body}\end{sideways}
   & \begin{sideways}{boat}\end{sideways}
   & \begin{sideways}{\textbf{average}}\end{sideways} \\
\hline
  MIM\cite{vezhnevets2011weakly}
  & 12 & 83 & 70 & \textbf{81} & \textbf{93} & 84 & \textbf{91} & 55 & \textbf{97} & \textbf{87} & 92 & 82 & 69 & \textbf{51} & 61 & 59 & 66 & 53 & 44 & 9 & \textbf{58} & 67  \\
\hline
   K. Zh\cite{zhang2013sparse} & \textbf{63} & 93 & \textbf{92} & 62 & 75 & 78 & 79 & 64 & 95 & 79 & \textbf{93} & 62 & 76 & 32 & \textbf{95} & 48 & \textbf{83} & 63 & 38 & \textbf{68} & 15 & 69  \\
\hline
   \textbf{Ours} & 45 & 73 & 65 & 79 & 81 & 66 & 71 & \textbf{87} & 75 & 84 & 73 & 73 & \textbf{94} & \textbf{51} & 89 & \textbf{85} & 42 & \textbf{83} & \textbf{81} & 66 & 32 & \textbf{71}  \\
\hline

\cline{1-23}
   \multicolumn{22}{| c |}{\textbf{VOC 2007}} \\
   \cline{1-22}
   \normalsize{Method}
   & \begin{sideways}{aeroplane}\end{sideways}
   & \begin{sideways}{bicycle}\end{sideways}
   & \begin{sideways}{bird}\end{sideways}
   & \begin{sideways}{boat}\end{sideways}
   & \begin{sideways}{bottle}\end{sideways}
   & \begin{sideways}{bus}\end{sideways}
   & \begin{sideways}{car}\end{sideways}
   & \begin{sideways}{cat}\end{sideways}
   & \begin{sideways}{chair}\end{sideways}
   & \begin{sideways}{cow}\end{sideways}
   & \begin{sideways}{diningtable}\end{sideways}
   & \begin{sideways}{dog}\end{sideways}
   & \begin{sideways}{horse}\end{sideways}
   & \begin{sideways}{motorbike}\end{sideways}
   & \begin{sideways}{person}\end{sideways}
   & \begin{sideways}{pottedplant}\end{sideways}
   & \begin{sideways}{sheep}\end{sideways}
   & \begin{sideways}{sofa}\end{sideways}
   & \begin{sideways}{train}\end{sideways}
   & \begin{sideways}{tvmonitor}\end{sideways}
   & \begin{sideways}{\textbf{average}}\end{sideways} \\
\cline{1-22}
  Shotton,weakly\cite{shotton2008semantic}
  & 14 & 8 & 11 & 0 & \textbf{17} & \textbf{46} & 5 & \textbf{13} & 4 & 0 & \textbf{30} & 29 & 12 & \textbf{18} & 40 & 6 & 17 & 17 & 14 & 9 & 16  \\
\cline{1-22}
  K. Zh\cite{zhang2013sparse}
  & 48 & \textbf{20} & \textbf{26} & \textbf{25} & 3 & 7 & \textbf{23} & \textbf{13} & \textbf{38} & \textbf{19} & 15 & 39 & 17 & \textbf{18} & 25 & \textbf{47} & 9 & \textbf{41} & 17 & 33 & 24  \\
\cline{1-22}
  \textbf{Ours} & \textbf{68} & 14 & 12 & 16 & 4 & 27 & 18 & 12 & 28 & 16 & 7 & \textbf{46} & \textbf{36} & 11 & \textbf{78} & 18 & \textbf{29} & 11 & \textbf{47} & \textbf{41}  & \textbf{27} \\
\cline{1-22}

\end{tabular}
\end{center}
\caption{Accuracies (\%) of our method for each category on MSRC and VOC 2007 dataset, in comparison with other algorithms. The last column is the average accuracy over all categories.}
\label{tb:msrc_and_voc}
\end{table*}

\begin{figure*}[t]
\begin{center}
\includegraphics[width=1\linewidth]{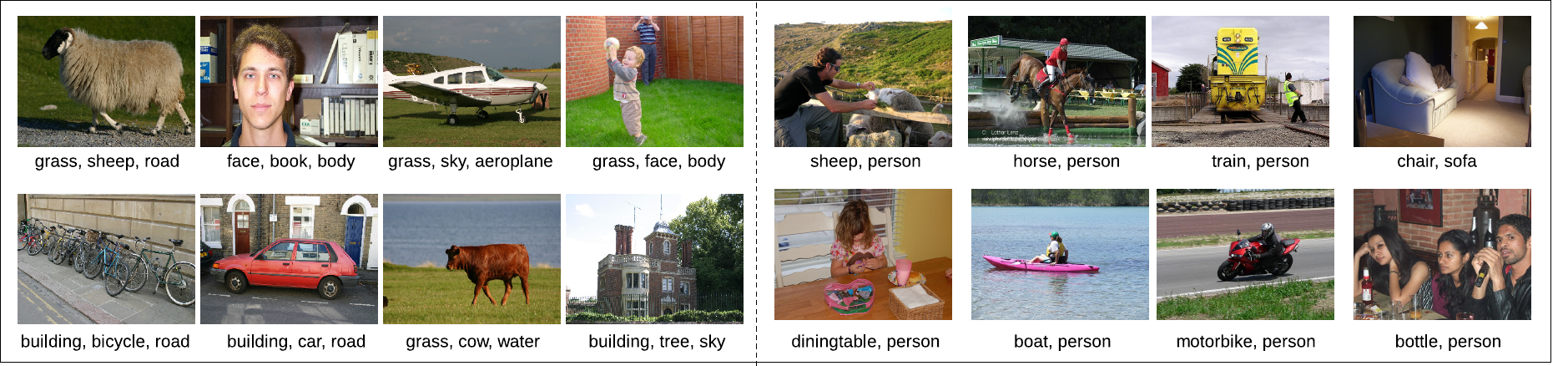}
\end{center}
\vspace{-5mm}
\caption{\small Some example results on image annotation from the MSRC (\textbf{left}) and VOC 2007 dataset (\textbf{right}).}
\label{fig_annotation}
\end{figure*}

\section{Experiment} \label{sec_exper}
In this section, we conduct extensive experiments to validate the performance of our method and discuss the experimental analysis. We also conduct an empirical study on the effectiveness of the proposed EM iterations.

\textbf{Implemenation details}:
Five parameters are required to be set in our framework. 
We set $q=20$ to construct the $q$-nearst graph, and set $p=10$ to retrieval $10$ images as reference for each test image. In the experiment we also set $\lambda=1$ empirically. The other parameters $\beta$ and $\gamma$ are introduced in Sec. (\ref{sec_pars}).


\begin{figure}[htb]
 \begin{center}
 \includegraphics[width=1\linewidth]{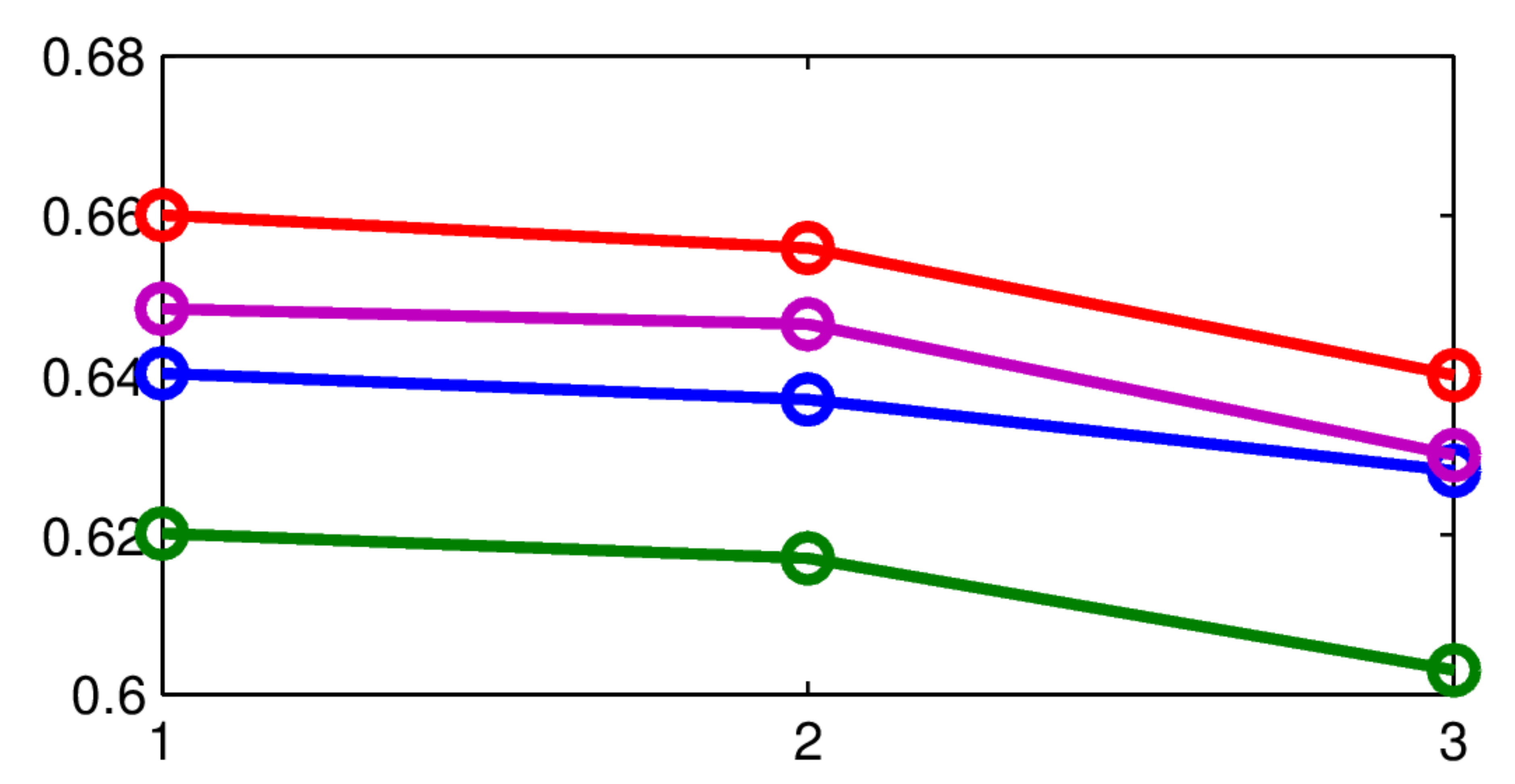}
 \end{center}
 \vspace{-5mm}
    \caption{Illustration of the decrease energy $E_{\alpha}$ decrease w.r.t. time. $x-axis$ indicates the number of iteration, and the $y-axis$ shows the energy $E_{\alpha}$ of Eq. (\ref{eq_combine}). The results randomly selected from test set.}
 \label{fig:energy_descend}
\end{figure}

\subsection{Datasets}
To verify the effectivenes of our method, we conduct experiments on two challenging datasets, i.e.  \textbf{MSRC}~\cite{shotton2006textonboost} and \textbf{VOC 2007} \cite{pascal-voc-2007}, by comparing with state-of-the-art. 
We use the standard average per-class measure (\emph{average accuracy}) to evaluate the performance. 
For each test image, we use the training set as the auxiliary data for our framework.



\subsection{Exp-I: Image Semantic Segmentation}

\subsubsection{Parameter Analysis} \label{sec_pars}
Specifically, we focus on the effects of $\beta$ and $\gamma$ which control the influence of appearance term and semantic term in Eq. (\ref{eq_combine}), and these two parameters are crucial to our results. The range of $\beta$ and $\gamma$ are both set to $\{0, 0.05, 0.10, 0.15, 0.20, 0.25, 0.30\}$. The semantic segmentation performance is used to tune parameters.

\begin{figure}[!htp]
\begin{center}
\includegraphics[width=1\linewidth]{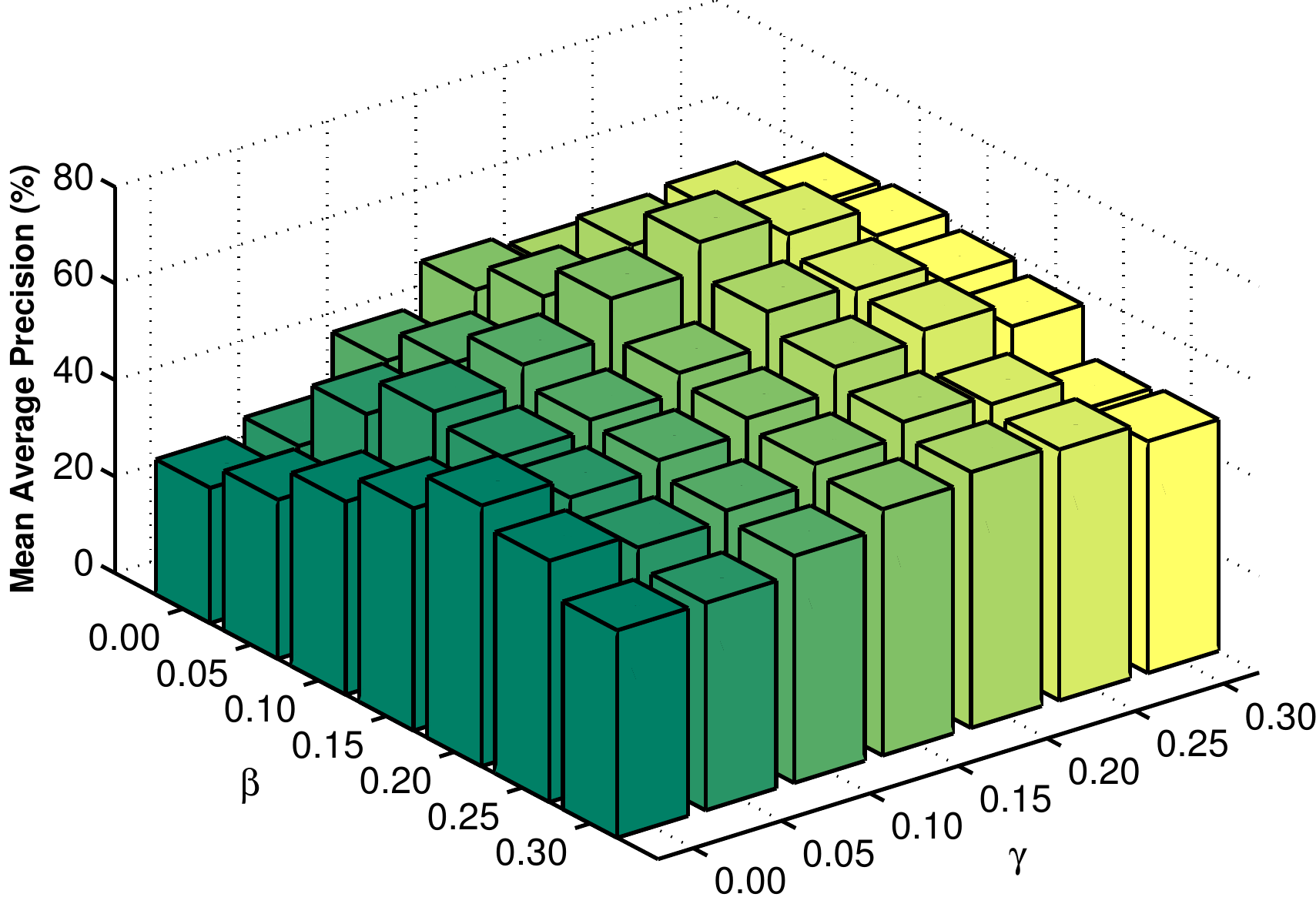}
\end{center}
\vspace{-5mm}
\caption{Parameter tuning results of parameters $\beta$ and $\gamma$ for \textbf{MSRC} dataset.}
\label{fig_pars}
\end{figure}

We used \textbf{MSRC} dataset to finetune the parameters. The results of changing the parameter values are presented in Fig. \ref{fig_pars}, from which we can observe the following conclusions:


\begin{itemize}
\item When $\beta$ and $\gamma$ increase from small values to large values, the performance varies apparently, which shows that the sparse term and semantic constraint term have great impacts on the performance.


\item Mean average precision (MAP) reach the peak points (0.71) when $\beta=0.1$ and $\gamma=0.2$ on \textbf{MSRC} which lie in the middle range and the precision do not increase monotonically when $\beta$ and $\gamma$ increase. 
    In the following experiments, we adopt the best parameter settings on all datasets.
\end{itemize}

\subsubsection{Experiments on MSRC dataset}
Given this insight, we compare the proposed method with the following stae-of-the-art algorithms: 
\textit{MIM}\cite{vezhnevets2011weakly}, and \textit{K. Zh}\cite{zhang2013sparse}. 

Table \ref{tb:msrc_and_voc} 
shows that our algorithm outperforms the others. Benefit from the semantic constraints incorporated in our approach, we achieve a significant improvements for certain difficult classes, \emph{e.g.},  \emph{chair} and \emph{cat}.  Serveral visualized results with the corresponding ground-truths are presented in Fig. \ref{fig:segment result}, and more semantic segmentation results are in supplementary material as to the limited space of article.

\subsubsection{Experiments on VOC 2007 dataset}
Few performance on VOC 2007 dataset is reported, due to the 20 extremely challenging categories it contains. 
Here we compare with the weakly supervised STF\cite{shotton2008semantic} by running the code provide by the author. We also compare our method with \cite{zhang2013sparse}. Results are reported in  Table \ref{tb:msrc_and_voc}, and our methods outperforms \cite{zhang2013sparse} by 3\%.

It takes about 8 seconds per image with an un-optimized matlab implementation for semantic segmentation, on a 64-bit system with Core-4 3.6 GHz CPU, 4GB memory (extracting features: 1s; sparse coding with semantic constraints: 5s; optimization by GraphCuts: 2s).

Moreover, we validate the effectiveness of the proposed EM iterations from two aspects. First, we plot the energy $E_{\alpha}$ in each iteration, which is the energy of semantic-aware spare coding defined in Eq. (\ref{eq_combine}), as shown in Fig. \ref{fig:energy_descend}. 
We also present some intermediate results during the EM iterations\footnote{Generally, the iteration is complete after two or three steps since the average number of labels for each image is 3 in MSRC or VOC2007 dataset.}, as Fig. \ref{fig:iteration_result} shown, which empirically supports the effectiveness of the iterations.


\subsection{Exp-II: Image Annotation on Test Image}
\subsubsection{Benchmarks and Metrics}
Three popular algorithms are implemented as benchmark baselines for the image annotation task:
\textbf{MAHR}\cite{huang2012multi-hypothesis}, \textbf{MLkNN}\cite{zhang2007ml}, \textbf{ML-LOC}\cite{huang2012multi}.

{MLkNN} and {ML-LOC} are the state-of-the-art multi-label annotation algorithms in literature. They have been reported to outperform most other multi-label annotating algorithms, such as RankSVM \cite{elisseeff2001kernel}. Thus, we do not plan to further implement the latter two in this work. We evaluate and compare among the three algorithms over two datasets, MSRC and VOC 2007, each of which is randomly and evenly split into training and testing subset. The image annotation performance is measured by mean average precision, which is widely used for evaluating the performances of ranking related tasks.

\subsubsection{Results and Analysis} \label{sec_annotation}
The \textbf{weighed} method is outperforms the unweighed one as Table \ref{tb:annotation} shown. It notices that the sparse coefficient $\boldsymbol{\alpha}$ is useful to improve the image annotation performance, and useful for image semantic segmentation apparently, as we do the image retrieval by jointly matching their appearance as well as the semantics. The larger $\boldsymbol{\alpha}_i$ means the more similar in semantics between the test image and image $I_i$ (i.e. sharing the more common labels).

The \textbf{weighed} method proposed outperforms the three classical methods listed in Table \ref{tb:annotation}. Some example image annotation results from the MSRC and VOC 2007 dataset are shown in Fig. \ref{fig_annotation}. Here we only display the top $3$ or $2$ labels for MSRC and VOC 2007, since the average number of labels for each images in MSRC and VOC 2007 is $3$ and $2$ respectively.

\begin{table}[htp]
\begin{center} \footnotesize
\begin{tabular}{|c|c|c|c|c|c|}
\hline
Dataset & MAHR & MLkNN  & ML-LOC & unweighed & \textbf{weighed} \\
\hline
MSRC & 49.5 & 70.8  & 77.3 &  76.1 &\textbf{84.7} \\
\hline
VOC 2007 & 34.0 & 47.6   & 48.9 & 45.8 &   \textbf{57.5}\\
\hline

\end{tabular}
\end{center}
\caption{Image label annotation MAP (Mean Average Precision) comparisons on  two different datasets.}

\label{tb:annotation}
\end{table}

\section{Conclusions} \label{sec_conclusion}

In this paper proposes a new framework for data-driven semantic image segmentation where only image-level labels are available, and it is also useful for image annotation. Compared with the traditional supervised learning methods, our framework is more flexible for real applications such as online image retrieval. In the experiments, we demonstrate very promising results on the standard benchmarks of scene understanding. In future work, we can improve the algorithm efficiency by utilizing parallel implementation and validate our approach on larger scale datasets.



\bibliographystyle{abbrv} 
\bibliography{ref}

\begin{thebibliography}{10}

\bibitem{elisseeff2001kernel}
A.~Elisseeff and J.~Weston.
\newblock A kernel method for multi-labelled classification.
\newblock In {\em NIPS}, volume~14, pages 681--687, 2001.

\bibitem{pascal-voc-2007}
M.~Everingham, L.~Van~Gool, C.~K.~I. Williams, J.~Winn, and A.~Zisserman.
\newblock The {PASCAL} {V}isual {O}bject {C}lasses {C}hallenge 2007 {(VOC2007)}
  {R}esults.
\newblock
  http://www.pascal-network.org/challenges/VOC/voc2007/workshop/index.html.

\bibitem{huang2012multi-hypothesis}
S.-J. Huang, Y.~Yu, and Z.-H. Zhou.
\newblock Multi-label hypothesis reuse.
\newblock In {\em Proceedings of the 18th ACM SIGKDD international conference
  on Knowledge discovery and data mining}, pages 525--533. ACM, 2012.

\bibitem{huang2012multi}
S.-J. Huang, Z.-H. Zhou, and Z.~Zhou.
\newblock Multi-label learning by exploiting label correlations locally.
\newblock In {\em AAAI}, 2012.

\bibitem{ladicky2010graph}
L.~Ladicky, C.~Russell, P.~Kohli, and P.~H.~S. Torr.
\newblock Graph cut based inference with co-occurrence statistics.
\newblock In {\em ECCV}. Springer, 2010.

\bibitem{lin2010layered}
L.~Lin, X.~Liu, and S.-C. Zhu.
\newblock Layered graph matching with composite cluster sampling.
\newblock {\em Pattern Analysis and Machine Intelligence, IEEE Transactions
  on}, 32(8):1426--1442, 2010.

\bibitem{lin2012representing}
L.~Lin, P.~Luo, X.~Chen, and K.~Zeng.
\newblock Representing and recognizing objects with massive local image
  patches.
\newblock {\em Pattern Recognition}, 45(1):231--240, 2012.

\bibitem{lin2014AOG}
L.~Lin, X.~Wang, W.~Yang, and J.-H. Lai.
\newblock Discriminatively trained and-or graph models for object shape
  detection.
\newblock {\em IEEE Transactions on Pattern Analysis and Machine Intelligence},
  2014.

\bibitem{lin2009grammar}
L.~Lin, T.~Wu, J.~Porway, and Z.~Xu.
\newblock A stochastic graph grammar for compositional object representation
  and recognition.
\newblock {\em Pattern Recognition}, 42(7):1297--1307, 2009.

\bibitem{liu2009nonparametric}
C.~Liu, J.~Yuen, and A.~Torralba.
\newblock Nonparametric scene parsing: Label transfer via dense scene
  alignment.
\newblock In {\em CVPR}. IEEE, 2009.

\bibitem{luo2012joint}
P.~Luo, X.~Wang, L.~Lin, and X.~Tang.
\newblock Joint semantic segmentation by searching for compatible-competitive
  references.
\newblock In {\em Proceedings of the 20th ACM international conference on
  Multimedia}, pages 777--780. ACM, 2012.

\bibitem{neal1998view}
R.~M. Neal and G.~E. Hinton.
\newblock A view of the em algorithm that justifies incremental, sparse, and
  other variants.
\newblock In {\em Learning in graphical models}. Springer, 1998.

\bibitem{shotton2008semantic}
J.~Shotton, M.~Johnson, and R.~Cipolla.
\newblock Semantic texton forests for image categorization and segmentation.
\newblock In {\em CVPR}. IEEE, 2008.

\bibitem{shotton2006textonboost}
J.~Shotton, J.~Winn, C.~Rother, and A.~Criminisi.
\newblock Textonboost: Joint appearance, shape and context modeling for
  multi-class object recognition and segmentation.
\newblock In {\em ECCV}. Springer, 2006.

\bibitem{vezhnevets2011weakly}
A.~Vezhnevets, V.~Ferrari, and J.~M. Buhmann.
\newblock Weakly supervised semantic segmentation with a multi-image model.
\newblock In {\em ICCV}. IEEE, 2011.

\bibitem{winn2005locus}
J.~Winn and N.~Jojic.
\newblock Locus: Learning object classes with unsupervised segmentation.
\newblock In {\em ICCV}. IEEE, 2005.

\bibitem{zhang2013sparse}
K.~Zhang, W.~Zhang, Y.~Zheng, and X.~Xue.
\newblock Sparse reconstruction for weakly supervised semantic segmentation.
\newblock In {\em IJCAI}. AAAI Press, 2013.

\bibitem{zhang2007ml}
M.-L. Zhang and Z.-H. Zhou.
\newblock Ml-knn: A lazy learning approach to multi-label learning.
\newblock {\em Pattern recognition}, 40(7):2038--2048, 2007.

\end{thebibliography}

\begin{IEEEbiography}[{\includegraphics[width=1in,height=1.25in,clip]{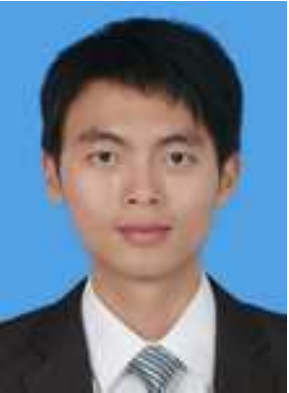}}]
{Xionghao Liu}
is currently a graduate student majored in Pattern Recognition and Computer Vision from Sun Yan-sen University, and has received his B.E. degrees in the School of Information Science and Technology, Sun Yat-sen University, Guangzhou, P. R. China, in 2012.
\end{IEEEbiography}

\vspace{-150 mm}
\begin{IEEEbiography}[{\includegraphics[width=1in,height=1.25in,clip]{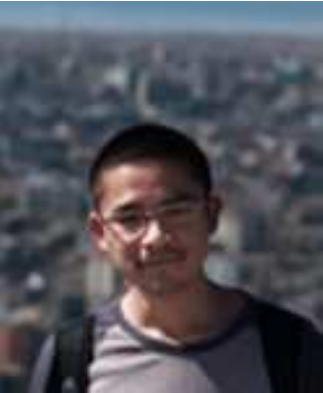}}]
{Wei Yang}
received his B.E. degree in Software Engineering, Sun Yat-sen University. He received his master degree in the Department of Computer Science, Sun Yat-sen University. He is currently a PhD student in the Department of Electronic Engineering, The Chinese University of Hong Kong. His research interests include computer vision and machine learning.
\end{IEEEbiography}

\vspace{-150 mm}
\begin{IEEEbiography}[{\includegraphics[width=1in,height=1.25in,clip]{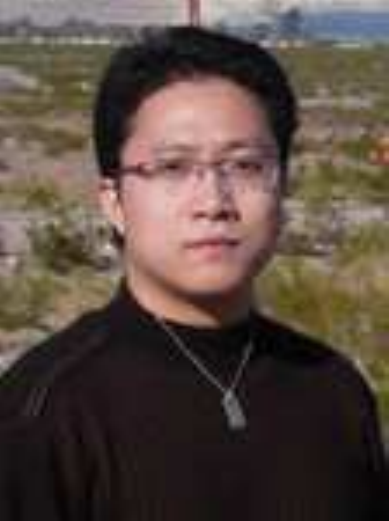}}]
{Liang Lin}
is a Professor with the School of Advanced Computing, Sun Yat-Sen University, China. His research focuses on new models, algorithms and systems for intelligent processing and understanding of visual data. He has published more than 60 papers in top tier academic journals and conferences, and has served as an associate editor for journal Neurocomputing and The Visual Computer.
\end{IEEEbiography}

\newpage

\begin{IEEEbiography}[{\includegraphics[width=1in,height=1.25in,clip]{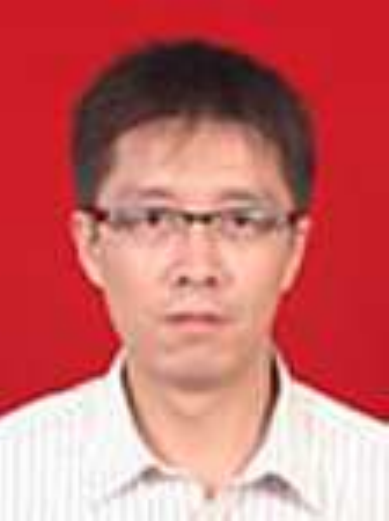}}]
{Qing Wang}
is an Associate Professor of Sun Yat-Sen University, China, Ph.D. in Computer Science and Member of SIGCHI, researcher on Human Computer Interaction, User Experience, Collaborative Software, and Web Usability, and especially interested in utilizing browser history on collaboration.
\end{IEEEbiography}

\vspace{-150 mm}
\begin{IEEEbiography}[{\includegraphics[width=1in,height=1.25in,clip]{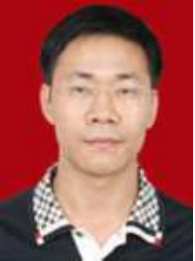}}]
{ZhaoQuan Cai}
was born in 1970, and is now a professor in Huizhou University, China. His research interest include computer networks, intelligent computing, and database systems.
\end{IEEEbiography}

\vspace{-150 mm}
\begin{IEEEbiography}[{\includegraphics[width=1in,height=1.25in,clip]{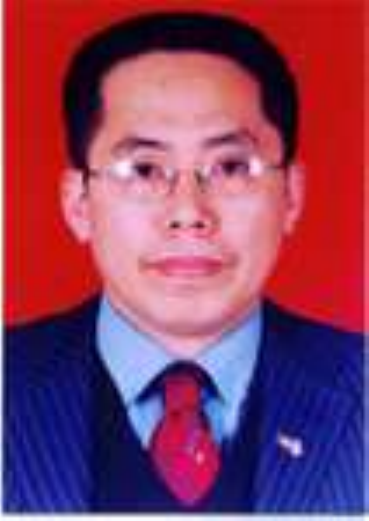}}]
{Jianhuang Lai}
is currently a Professor and the Dean of the School of Information Science and Technology. His research focuses on image processing, pattern recognition, multimedia communication, wavelet, and its applications. He serves as a Standing Member of the Image and Graphics Association of China and a Standing Director of the Image and Graphics Association of Guangdong.
\end{IEEEbiography}

\end{document}